\documentclass{article}
\usepackage{PRIMEarxiv}

\usepackage[utf8]{inputenc}      
\usepackage[T1]{fontenc}         
\usepackage[hidelinks]{hyperref}

\usepackage{amsmath}
\usepackage{amssymb}
\usepackage{amsfonts}
\usepackage{microtype}      

\usepackage{dsfont}
\usepackage{enumitem}
\usepackage{float}          
\usepackage{graphicx}
\graphicspath{{fig/}}       

\usepackage{algorithm}
\usepackage{algorithmicx}
\usepackage[noend]{algpseudocode}

\usepackage{booktabs}
\usepackage{array}
\newcolumntype{C}[1]{>{\centering\arraybackslash}p{#1}}
\newcolumntype{R}[1]{>{\raggedleft\arraybackslash}p{#1}}

\usepackage[T1]{fontenc}
\usepackage{inconsolata}
\usepackage{listings}
\usepackage{caption}
\usepackage{xcolor, soul}
\DeclareCaptionFormat{listing}{#1#2#3}
\lstdefinestyle{python}{
    language=Python,
    basicstyle=\ttfamily\small,
    keywordstyle=\color{blue},
    stringstyle=\color{orange},
    commentstyle=\color[rgb]{0.13, 0.55, 0.13},
    frame=single,
    showstringspaces=false,
    breaklines=true,
    morestring=[b][\color{orange}]{'''},
    morestring=[b][\color{orange}]{"""},
}

\usepackage{amsthm}
\newtheorem{theorem}{Claim}

\newcommand{\pbc}[1]{\left[ #1 \right]_{\circlearrowright}}

\newcommand{\Nc}{\mathcal{N}}

\newcommand{\Uc}{\mathcal{U}}

\newcommand{\fv}{\mathbf{f}}
\newcommand{\gv}{\mathbf{g}}

\newcommand{\xv}{\mathbf{x}}

\newcommand{\zv}{\mathbf{z}}

\newcommand{\Ev}{\mathbf{E}}

\newcommand{\epsilonv   }{\boldsymbol \epsilon   }

\newcommand{\muv        }{\boldsymbol \mu        }

\newcommand{\sigmav     }{\boldsymbol \sigma     }

\newcommand{\Thetav     }{\boldsymbol \Theta     }

\begin{document}
\label{sec:info}



\title{MDDM: A Molecular Dynamics Diffusion Model to Predict Particle Self-Assembly}


\author{
  Kevin Ferguson \qquad Yu-hsuan Chen \qquad Levent Burak Kara\\
  Department of Mechanical Engineering \\
  Carnegie Mellon University \\
  5000 Forbes Ave, Pittsburgh, PA 15213\\
  \texttt{lkara@andrew.cmu.edu}\\
}

\maketitle \vspace{-0.2in}


\begin{abstract}
    The discovery and study of new material systems rely on molecular simulations that often come with significant computational expense. We propose MDDM, a Molecular Dynamics Diffusion Model, which is capable of predicting a valid output conformation for a given input pair potential function. After training MDDM on a large dataset of molecular dynamics self-assembly results, the proposed model can convert uniform noise into a meaningful output particle structure corresponding to an arbitrary input potential. The model's architecture has domain-specific properties built-in, such as satisfying periodic boundaries and being invariant to translation. The model significantly outperforms the baseline point-cloud diffusion model for both unconditional and conditional generation tasks.
\end{abstract}



\section{Introduction}
\label{sec:intro}


Molecular Dynamics (MD) is a powerful computational tool that lets scientists and engineers study chemical, biological, or material systems at a micro- or nano-scale. We target a materials science application of molecular self-assembly in which the goal is to model the dynamics and structure of bulk systems containing many particles that interact with one another via a specified potential energy function. By simulating the motion and interaction of particles in a molecular system, material properties can be measured from the resulting equilibrated particle structures.  

While MD provides engineers with the capacity to perform high-fidelity material simulations, it is not without its own limitations, namely computational expense.  For one, to emulate the properties of a bulk material as accurately as possible, very large systems (i.e. systems with many particles) are required. Crucially, the number of particle-to-particle interactions in a system scales quadratically with the number of particles. While each time step's computations can be parallelized across multiple processors, the time-evolving nature of an MD simulation is inherently serial, leading to prohibitively lengthy simulation times, especially for many-particle systems.

Predicting the final set of particle locations in an MD simulation, often referred to as conformation generation \cite{Hawkins_2017}, is a popular area of interest in molecular research, which has been approached from a number of machine learning-based directions. Spellings et al.~\cite{Spellings_2018} characterize self-assembled molecular structures by classifying their phase, leveraging Gaussian Mixture Models and various supervised learning techniques. While effective, they do not address the \textit{generative} problem associated with creating a valid crystal structure. As for generative models, Arts et al.~\cite{Arts_2023} use a diffusion model approach to predict the free energy coarse-grained MD structures, while Wu et al.~\cite{Wu_2023} generate the steady-state conformation of molecules given their connectivity graphs. These approaches are similar to ours, but fail to solve a large-scale self-assembly generation task, as their simulations contain only a few dozen particles, rather than hundreds/thousands. Some authors attempt the inverse problem, trying to recover a suitable potential for a given output simulation \cite{designing-isotropic-2011, novel-self-assembled-2011, inverse-design-2018}. However, these are iterative methods that require performing simulations in-the-loop. A generative surrogate would significantly accelerate such processes. GeoDiff, a diffusion model proposed by Xu et al.~\cite{geodiff-xu-2022}, features a custom roto-translational invariant graph convolution in a diffusion framework to generate conformations of molecules. Our work is similar, but it differs in that we extend the theory of denoising diffusion such that periodic boundary conditions can be satisfied; we also target large-scale material simulation rather than structure generation for molecules.

While the specific problem we pose has not been tackled using generative AI, generative denoising diffusion models in the point-cloud domain have indeed been recently explored in a number of areas, such as high-energy physics simulation \cite{fast-mikuni-2023}, anatomical reconstruction \cite{implant-friedrich-2023}, and protein backbone generation \cite{se3-yim-2023}. Point-cloud diffusion models can also generate point clouds of common objects, having been trained on their respective shape datasets \cite{point-cloud-diffusion-2021, lion-zheng-2022}. Earlier, Generative Adversarial Networks (GANs) were used for the same purpose \cite{point-cloud-gan-2018}. These methods make use of point cloud encoders such as PointNet \cite{pointnet-2017}. Other methods such as graph networks \cite{edgeconv-2019} or transformer-based approaches \cite{transolver-2024} can be used for this intermediate task as well.

\begin{figure}[t]
    \centering
    \includegraphics[width=1.0\textwidth]{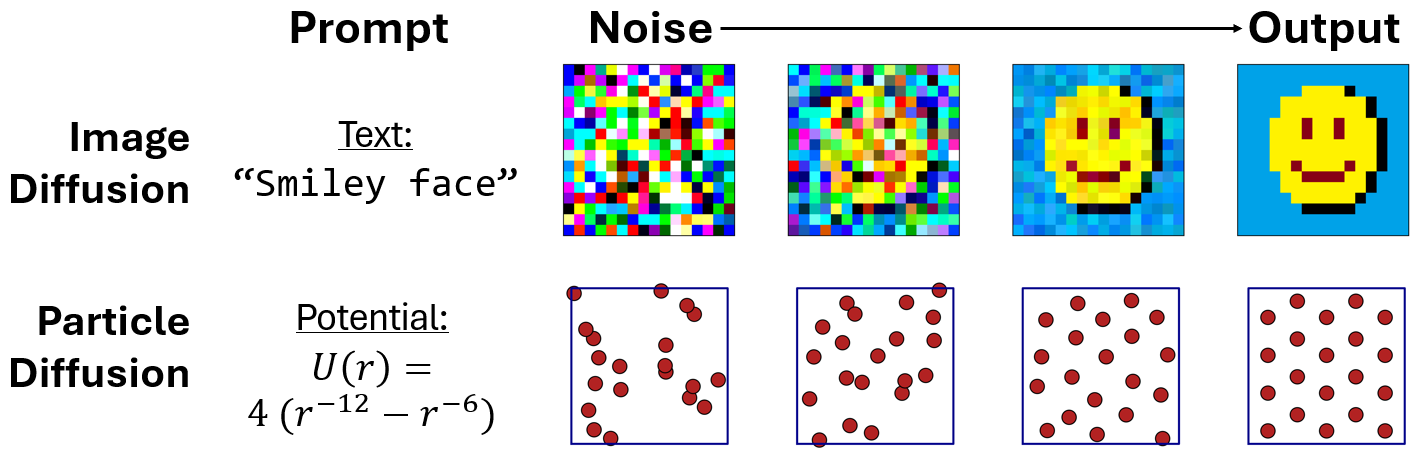}
    \caption{Text-conditioned image diffusion models, compared with the proposed potential-conditioned particle diffusion model for MD self-assembly conformation prediction}
    \label{fig:particle-diffusion}
\end{figure}

In this work, we propose a denoising diffusion probabilistic model (i.e., a diffusion model) that can act as a surrogate model for large-scale MD. A diffusion model starts with noise, and -- through an iterative denoising process -- generates a sample from the dataset distribution. Most often, diffusion models are image-generators, producing realistic images, often conditioned on a text prompt. In our proposed Molecular Dynamics Diffusion Model (MDDM), rather than creating an image from an input prompt, a set of particle locations is created from an input pair potential function. This distinction is illustrated in Fig.~\ref{fig:particle-diffusion}.

We demonstrate both unconditional and conditional conformation generation for particle self-assembly. MDDM properly accounts for the periodic boundary conditions present in MD. Our model generates structures that match the target radial distribution functions qualitatively and quantitatively better than the baseline. By sampling from the MDDM model, the structure of materials systems can be obtained rapidly, with far fewer diffusion model denoising iterations (hundreds) compared to MD time steps (millions), letting an engineer study bulk material systems more efficiently than ever before.



Our key contributions can be summarized  as follows:

\begin{enumerate}
    \item A large dataset of MD simulation results spanning a range of input system potentials and temperatures
    \item MDDM, a Molecular Dynamics Diffusion Model that can generate a self-assembled structure from an input pairwise potential energy function
    \item A periodic boundary graph network model suitable for denoising a particle configuration in a molecular dynamics context
\end{enumerate}


\section{Methods}
\label{sec:methods}


In this section, we formalize the self-assembly conformation generation task we aim to solve using a diffusion-based approach. We also describe the dataset, define the proposed MDDM model, and outline procedures for training and sampling from the model.


\subsection{Dataset}
\label{sec:dataset}

We have generated a large dataset of MD results for this problem using LAMMPS \cite{LAMMPS} because there is not an existing dataset for this task. The simulations consist of 1000 particles each within a $10\times10\times10$ box (units are dimensionless `LJ' units). The system is annealed to the target temperature from a temperature $10\times$ larger, and the pair potential is an \textit{oscillating pair potential} (OPP) given by $U_{\text{OPP}}(r) = r^{-15} + r^{-3}\cos(k(r-1.25)-\phi)$, for parameters $k$ and $\phi$. This potential was introduced by Mihalkovi\v{c} and Henley \cite{opp-mihalkovic-2012} as a concise potential form capable of producing the complex behavior of quasi-crystal structures. The inputs and outputs of the simulation are summarized as follows:

\begin{itemize}[leftmargin=2cm]
    \item[\textbf{Inputs:}] Potential frequency parameter $k$; Potential shift parameter $\phi$; Target system temperature $T$
    \item[\textbf{Outputs:}] Final particle locations $\boldsymbol{X}_{(N\times 3)}$
\end{itemize}

We performed simulations across the parameter ranges: $1.0 \leq k \leq 15.0$;\; $0.0 \leq \phi \leq 6.0$;\; $0.01 \leq T \leq 0.05$, with 10 values across each input parameter, for a total of 1000 initial simulations. Visualizations of a few of these results are shown in Figure \ref{fig:gallery} in the Appendix.


\subsection{Task and Benchmark}
\label{sec:task}

\begin{figure}[t]
    \centering
    \includegraphics[width=1.0\textwidth]{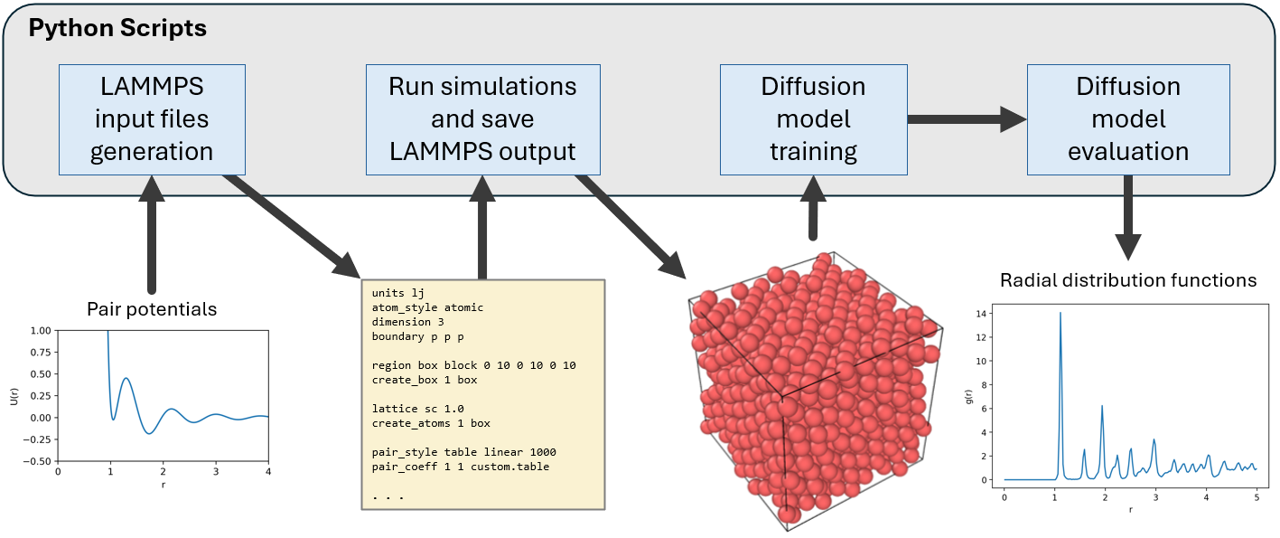}
    \caption{The dataset generation, model training, and model evaluation steps toward the MDDM task.}
    \label{fig:plan}
\end{figure}

The final goal is to, given the MD inputs from above, generate a set of particle locations that is thermodynamically indistinguishable from an MD trajectory output that had the same inputs. In probabilistic modeling terms, the goal of MDDM will be to sample from the distribution of all possible output MD structures that correspond to a given input potential and temperature. The necessary components are summarized in Fig.~\ref{fig:plan}, which illustrates the inputs and outputs, as well as the flow of modular scripts we have implemented to perform each step.

For an evaluation metric, we wish to compare the simulation structure output with those generated via our model. The radial distribution function (RDF) is a useful function that describes the normalized density of a particle system, with respect to a reference particle. We will also include RDF visualizations for comparison. RDFs are often visually compared to ensure no significant mismatch, as in \cite{graph-accelerated-2022}. This is especially important because visual inspection of particle structures themselves is infeasible, as is clear from Figure \ref{fig:gallery}. For a quantitative result to allow comparison between the results of multiple distinct methods, we use a custom loss function RDF-MSE, which computes the mean-squared-error between two RDF curves. Specifically, the RDF is computed as a normalized histogram of particle-to-particle distances with $n_b$ bins spanning from 0 to half of the simulation box side length; the RDF-MSE thus compares two $n_b$-element vectors, where $n_b=100$ has been selected as a hyperparameter.


Due to of the lack of generative models for material self-assembly in the MD domain, we compare against a somewhat disparate baseline: Diffusion with Transolver. That is, we use Transolver \cite{transolver-2024} as a means of iteratively denoising a point cloud from noise to structure. The node-embedding capabilities of Transolver allow it to process an input shape and predict a suitable noise vector at each node. We apply the same diffusion strategy for the baseline as we do for our proposed model, which was inspired in part by Diffusion Point Cloud \cite{point-cloud-diffusion-2021}.


\subsection{Model Architecture}

Our approach is a denoising diffusion probabilistic model that operates in the 3-D point domain (rather than on 2-D images, as is typical of diffusion models). The input to our denoising process will be a set of particle coordinates sampled at random. These particle locations will be passed into a noise predictor model along with $k, \phi,$ and $T$ (the inputs to the MD system -- akin to the text prompt in an image diffusion model). The resulting prediction is a displacement of each particle, which we add to the initial positions to yield the denoised positions. This denoising procedure is repeated for $T$ iterations, and the final output is a particle conformation that corresponds to the inputs. 

\begin{figure}[t]
    \centering
    \includegraphics[width=\textwidth]{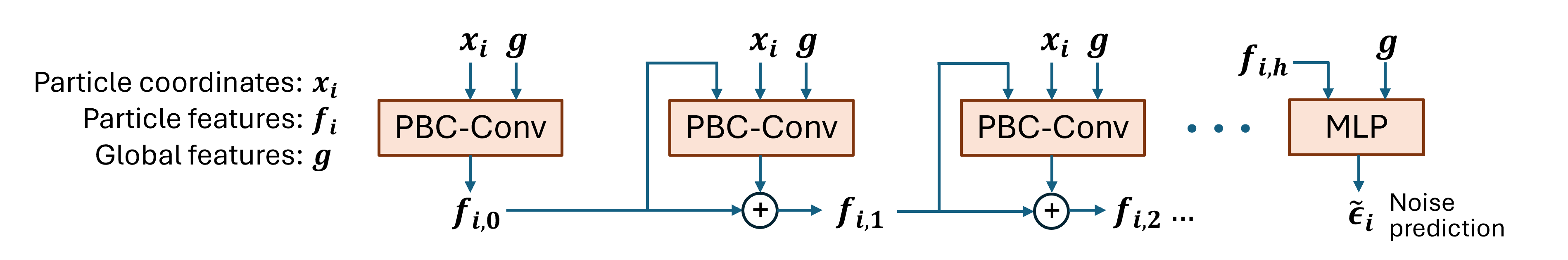}
    \caption{Denoising graph neural network architecture}
    \label{fig:denoiser}
\end{figure}

As previously discussed, the diffusion model architecture is defined in analogy with image diffusion methods, as shown in Fig.~\ref{fig:particle-diffusion}. The denoiser must therefore operate not on image data, but on a point cloud. Furthermore, our denoising step has been designed specifically to operate in the presence of periodic boundary conditions. Figure \ref{fig:denoiser} depicts the full denoising model architecture.

At each PBC-Conv layer, first a k-nearest neighbors graph is constructed, wrapping properly in the periodic domain. We implement a set of simple functions in PyTorch \cite{pytorch-2017} to generate this; code for a simplified implementation is included in \ref{sec:listings}. Once generated for a set of particles, the same graph can be re-used for subsequent PBC-Conv layers, until the denoising step has been applied, modifying particle positions. PBC-Conv layers are learnable convolution functions defined as follows:

\begin{equation} \label{eqn:pbc-conv}
    \fv_i' = \max_{j \in \Nc(i)} h_{\boldsymbol{\Thetav}}(\;\pbc{\xv_j - \xv_i} \quad||\quad \fv_i \quad||\quad \fv_j - \fv_i \quad||\quad \gv\;),
\end{equation} 

where $\fv_i'$ is the output convolved set of features of particle $i$, $\Nc(i)$ are the periodic $k$-nearest neighbors of particle $i$, $\pbc{\xv_j - \xv_i}$ refers to the vector from neighbor $\xv_j$ to source $\xv_i$ (wrapped across periodic boundaries as needed), $\fv_i$ is the input pre-convolution set of features at particle $i$, and $\gv$ is the global feature vector; $\gv$ contains information about the MD system, such as potential parameters and temperature, as well as diffusion process parameters, like diffusion fraction $t/T$. (Note that $||$ denotes concatenation along the feature dimension.) This function is based on the EdgeConv operation from Wang et al.~\cite{edgeconv-2019}, but accounts for periodic boundaries and does not explicitly pass absolute coordinates into the model, making our model translation-invariant (also a property of MD simulations). Rotation-invariant graph convolutions have been proposed for atomic and molecular contexts like this \cite{geodiff-xu-2022, rotationinvariant-shuaibi-2021}, but we do not explicitly provide our network with this property. Rather, we augment our dataset with flipped, rotated, and axis-swapped data conformations, such that the network can implicitly learn the notion of rotation invariance as needed.

Our network consists of 8 PBC-Conv layers with hidden local/global feature counts of 32 at each layer, using a periodic $k$-nearest neighbors graph with $k=32$. Each PBC-Conv layer contains an MLP with 2 hidden layers of 32 neurons each. The output MLP has 2 layers, each with 128 neurons. For unconditional conformation generation, the sole global input was $t/T$, current diffusion time step divided by the total number of diffusion steps, with $T=500$ for our experiments; for conditional generation, temperature and potential parameters (each scaled within the range $[-1, 1]$) are set as additional global coordinates.


\subsection{Training and Sampling}

For a standard image-based DDPM, the forward diffusion step is given by: 
\begin{equation} \label{eqn:fwd-noise-standard}
    \xv_{t} = \sqrt{\alpha_t}\xv_0 + \sqrt{1 - \alpha_t}\epsilonv,\; \text{for}\; \epsilonv \sim \Nc(\boldsymbol{0}, \boldsymbol{I}),
\end{equation}
in which $\xv_{0}$ is the data, $\xv_{t}$ is the data corrupted by $t$ steps of diffusion, and $\alpha_t$ is a constant indicating the magnitude of diffusion from step $0$ to step $t$. However, use of this equation is predicated on the notion that at the final diffusion step, pixel values are normally distributed. For our systems, due to the presence of PBCs, particle locations will not obey a normal distribution after diffusion. Rather, they will take on a (very nearly) uniform distribution, since wrapping a normal distribution makes it become approximately uniform. A proof of this statement can be found in \ref{sec:proofs}. Because the final noise is therefore not normally distributed, the standard DDPM variational inference algorithm must be slightly reformulated. To do so, we redefine the forward diffusion process as follows:

\begin{equation} \label{eqn:fwd-noise}
    \xv_{t} = \text{wrap}\left(\xv_0 + \sqrt{\alpha_t}\epsilonv\right),\quad \text{for}\; \epsilonv \sim \Nc(\boldsymbol{0}, \boldsymbol{I}).
\end{equation}

Now, we can continue to use properties of Gaussian distributions, and maximization of the Evidence Lower Bound (ELBO) during training can be re-derived for this revised formulation. Note that the result in Eqn.~\ref{eqn:fwd-noise} is also wrapped within the periodic box domain. This gives rise to the training and sampling pseudocode in Algorithms \ref{alg:train} and \ref{alg:sample}. A probabilistic derivation of the mean and variance seen in the denoising step of Algorithm \ref{alg:sample} is provided in \ref{sec:proofs}.
For our experiments, we use $T = 500$ diffusion steps with $\alpha$ on a cosine-schedule $\alpha_t = \cos^2\left(\frac{\pi}{2}\cdot\frac{x/(T+1) + s}{1+s}\right)$, with $s = 0.008$. We train for 800 epochs using an Adam optimizer with learning rate 0.005, which decreases by a factor of 0.95 every 100 epochs.

\begin{figure}[H]
\begin{minipage}{0.49\linewidth}
\begin{algorithm}[H]
\caption{Training}\label{alg:train}
    \begin{algorithmic}[1]
        \Repeat
        \State $\xv_0 \sim q(\xv_0)$
        \State $t\sim \text{Uniform}(\{1,...,T\})$
        \State $\epsilonv \sim \mathcal{N}(\boldsymbol{0},\boldsymbol{I})$
        \State $\xv_t \gets \text{wrap}(\xv_0 + \sqrt{\alpha_t} \epsilonv)$ 
        \State Take optimizer step on $L_2$ loss, 
         $\nabla_\theta \| \epsilonv - \boldsymbol{\epsilon}_\theta (\xv_t, t) \|_2$
        \Until converged
    \end{algorithmic}
\end{algorithm}
\end{minipage}
\hfill
\begin{minipage}{0.49\linewidth}
\begin{algorithm}[H]
\caption{Sampling}\label{alg:sample}
    \begin{algorithmic}[1]
        \State $\xv_T \sim \mathcal{U}(\boldsymbol{0},\boldsymbol{L})$ \Comment{$L$ is box side length}
        \For {$t=T,...,1$}
            \State $\zv \sim \mathcal{N}(\boldsymbol{0},\boldsymbol{I}) \text{ if } t>1 \text{, else } \zv=0$ 
            \State $\muv_t \gets \left( \frac{\alpha_{t-1}}{\alpha_t}-1 \right)\cdot \left(\sqrt{\alpha_t}\; \boldsymbol{\epsilon}_\theta (\xv_t, t)\right) + \xv_t$ 
            \State $\sigmav^2_t \gets \frac{\alpha_{t-1}}{2\alpha_t} (\alpha_t - \alpha_{t-1})$         
            \State $\xv_{t-1} \gets \text{wrap}(\muv_t + \sigmav_t \zv)$ 
        \EndFor 
        \State \Return $\xv_0$
    \end{algorithmic}
\end{algorithm}
\end{minipage}
\end{figure}


\section{Experiments and Results}
\label{sec:experiments}


We run experiments on two versions of the MD task: unconditional generation and conditional generation. For unconditional generation, we train each model to generate a single randomly-selected MD conformation from our dataset -- no condition prompt is fed into the model. The conditional generative model, on the other hand, is trained on MD conformations from several potential parameter and temperature combinations, to enable generation of conformations that correspond to particular ``prompts'' of input temperature and potential parameter values. The datasets are augmented with rotated/flipped versions of this conformation to further impart translation/rotation invariance, and to avoid memorization of coordinate locations.

Training was carried out and performance was evaluated for two models: Diffusion with Transolver and MDDM. The Transolver model, with 128,867 parameters, achieved a training time of 24.03 seconds per 1,000 iterations and a sampling time of 2.53 seconds for 500 steps per data. In comparison, MDDM, with 101,891 PBC-Conv parameters, recorded a slightly faster training time of 23.15 seconds per 1,000 iterations but a longer sampling time of 4.17 seconds for 500 steps per data. These times are for a single NVIDIA RTX A6000 GPU with 128GB DDR4 RAM. The increased sampling time for our method can be partially attributed to the repeated graph generation, which requires calculating each particle's relative distance to its neighbors once at the start of each denoising step. This added computational complexity likely contributes to the observed difference in sampling efficiency.

For conditional generation, we train models to generate valid output conformations given $k$, $\phi$, and $T$ as additional input conditions, for the range of input parameters described in Sec.~\ref{sec:dataset}. To evaluate our diffusion model, we pass a set of MD inputs into the model and compare the resulting structures using the mean squared error (MSE) between their radial distribution functions (RDFs). We present results for both models on unconditional and conditional tasks, with average RDF-MSE values shown in Tab.~\ref{tab:results}.

\begin{figure}[p]
    \centering
    \includegraphics[width=\textwidth]{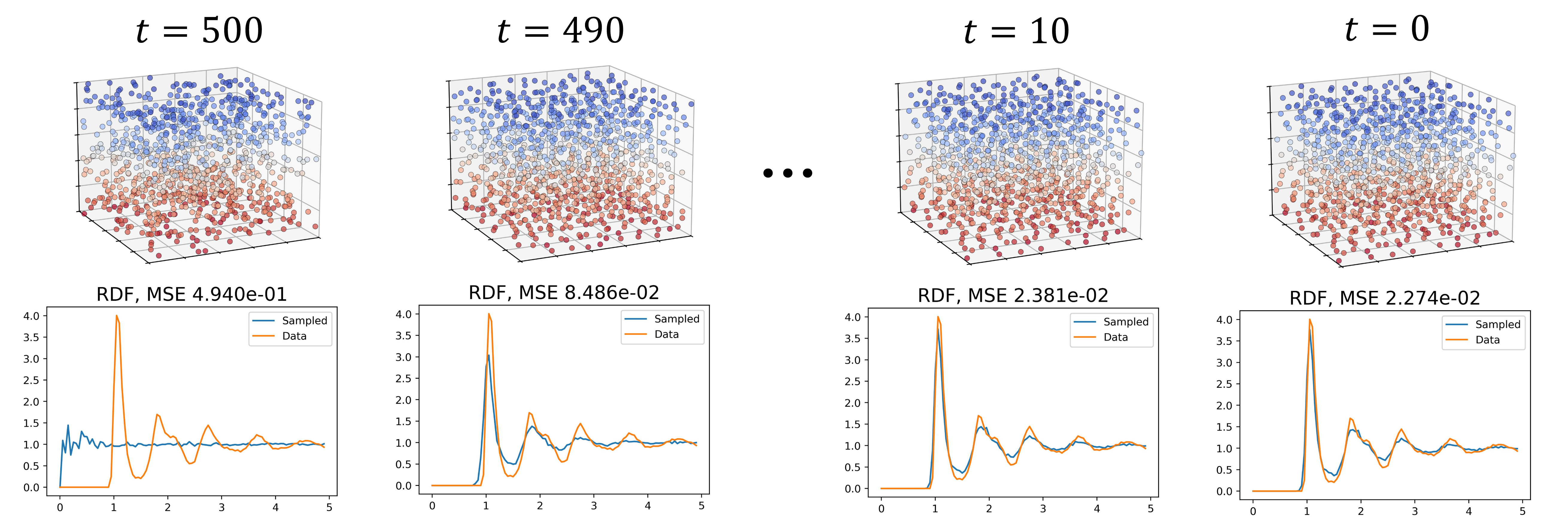}
    \caption{A demonstration of the sampling process for the unconditional generation task. The structure is mostly achieved in 10 diffusion steps, but RDF-MSE continues to decrease during subsequent steps}
    \label{fig:unconditional}
\end{figure}

For the unconditional task, the reverse noising process for sampling is visualized in Fig.~\ref{fig:unconditional}; by the end of the denoising process, at time step 0, we see a close match between the generated outputs and the ground truth -- this, along with the low MSE value in Tab.~\ref{tab:results}, indicates that the denoising procedure can indeed perform large-scale MD conformation generation. The baseline diffusion with Transolver model, however, struggles to yield performance as strong as our MDDM model, likely due to it prioritizing recreation of global shape information rather than local structure. We further observe that for MDDM, within only the first 10 of 500 steps, the RDF has nearly converged, with only small improvements occurring for subsequent sampling time steps, although the RDF-MSE score continues to decrease, indicating additional improvement. Ideally, there would be a more uniform transformation from the noise distribution to the data distribution. To achieve this in future work, we can explore implicit diffusion modeling \cite{ddim-song-2020}, alternative noise scheduling approaches \cite{improved-nichol-2021}, or diffusion-adjacent methods like flow-matching \cite{flow-lipman-2022}.

\begin{table}[p]
    \renewcommand{\arraystretch}{1.1}
    \centering
    \caption{RDF-MSE performance of our method compared to the baseline method}
\begin{tabular}{R{2.7cm}||C{2.5cm}||C{3cm}|C{3cm}} \hline
    \vspace{3pt}\textbf{Model} & \textbf{RDF-MSE Unconditional} & \textbf{RDF-MSE Conditional (Train)} & \textbf{RDF-MSE Conditional (Test)} \\ \hline
        Diff. w/ Transolver & 0.482 & 0.523 & 0.527 \\
        \textit{MDDM (Ours)} & \textbf{0.023} & \textbf{0.098} & \textbf{0.126} \\
        \hline \end{tabular}
    \label{tab:results}
\end{table}

\begin{figure}[p]
    \centering
    \includegraphics[width=0.63\textwidth]{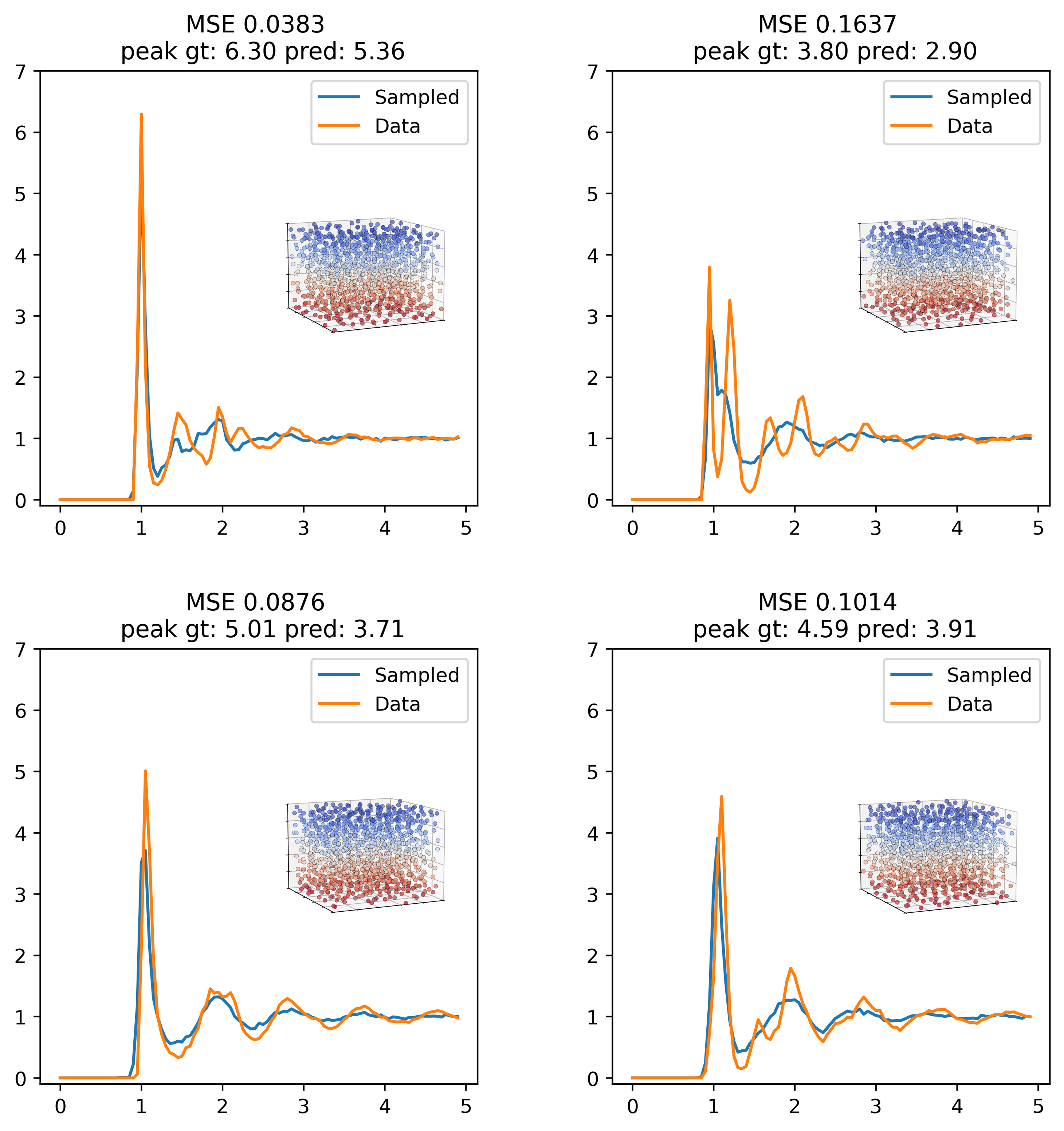}
    \caption{Four randomly selected results in the conditional generation task. Note the strong prediction of the first peak, followed by less accurate RDF match at larger length scales.}
    \label{fig:conditional}
\end{figure}


Figure \ref{fig:conditional} shows a small set of \textit{conditional} generation results on testing data. Here, the potential and temperature parameters were passed to the model as an additional set of node-wise input channels, now using all 1000 MD results as training data. Test data was generated by running an additional 25 simulations with new potential/temperature parameters uniformly randomly sampled from the training range. Diffusion with Transolver still struggles for this task, but MDDM matches the first RDF peak well. In an MD setting, the first peak is related to the coordination number of the RDF, which can correlate to the phase of the resulting material \cite{mechanics-chandler-1987, coordination-mikolaj-1968}. Therefore, even this initial model may have utility, e.g. as a material phase classifier.

Subsequent peaks in the RDF, however, exhibit some mismatch, so there is still nuance to the conditional task that our model struggles to capture. To address this, updating the diffusion framework to condition results using a cross-attention or guidance-based method is worth investigating. For example, Classifier Guidance \cite{guidance-dhariwal-2021} and Classifier-Free Guidance \cite{classifierfree-ho-2022} offer techniques for steering diffusion models toward high-quality conditioned results by ``guiding'' the denoising process. Methods like this are more sophisticated than the simple concatenation-based conditioning used in this work, and have greater potential to accurately model variations in potential and temperature.


\section{Conclusion}
\label{sec:conclusion}


We recontextualize diffusion models in a new domain: conformation generation for molecular dynamics material self-assembly. Through a redefinition of the forward noising process, our method enables potential-conditioned reverse diffusion of particles across periodic boundaries. 

The proposed MDDM model uses a custom periodic graph neural network to iteratively denoise uniformly distributed particles into meaningful structures. For training data, we generated a dataset of 1000 MD self-assembly results for different input potentials and temperatures. The trained model takes potential/temperature information as a condition, and it outputs a corresponding thermodynamically valid particle conformation. MDDM significantly outperforms point-cloud-based diffusion models for MD particle conformation generation. Unconditional generation has strong results, but there is still room for improvement in the conditional generation task.  

In future work, we hope to improve the results of conditional generation by leveraging alternative diffusion model conditioning methods. We also plan to investigate the effects of increasing the capacity of the model, as well as redesigning its architecture, or evaluating how the model can perform for MD materials systems with different underlying potentials. In particular, exploration of MD systems with non-isotropic and/or many-body potential functions would be another interesting extension of this work.



\section*{Acknowledgments}
This research was supported in part by the Center of Excellence established between the Air Force Research Laboratory and Carnegie Mellon University (AFOSR FA8650-19-5209). We thank the following individuals for their thoughtful discussion: Thomas O'Connor, Michael Bockstaller, Ayesha Abdullah, Andrew Gillman, Eric Harper, Jeffrey Ethier, Larry Drummy, Jacob Rast, and Matt Gormley.


\bibliographystyle{unsrt}  
\bibliography{references} 

\appendix
\renewcommand{\thefigure}{A\arabic{figure}}
\setcounter{figure}{0} 
\renewcommand{\thetable}{A\arabic{table}}
\setcounter{table}{0} 
\renewcommand{\thesection}{Appendix \Alph{section}}
\setcounter{section}{0}



\section{Dataset visualization}

\begin{figure}[H]
    \centering
    \includegraphics[width=\textwidth]{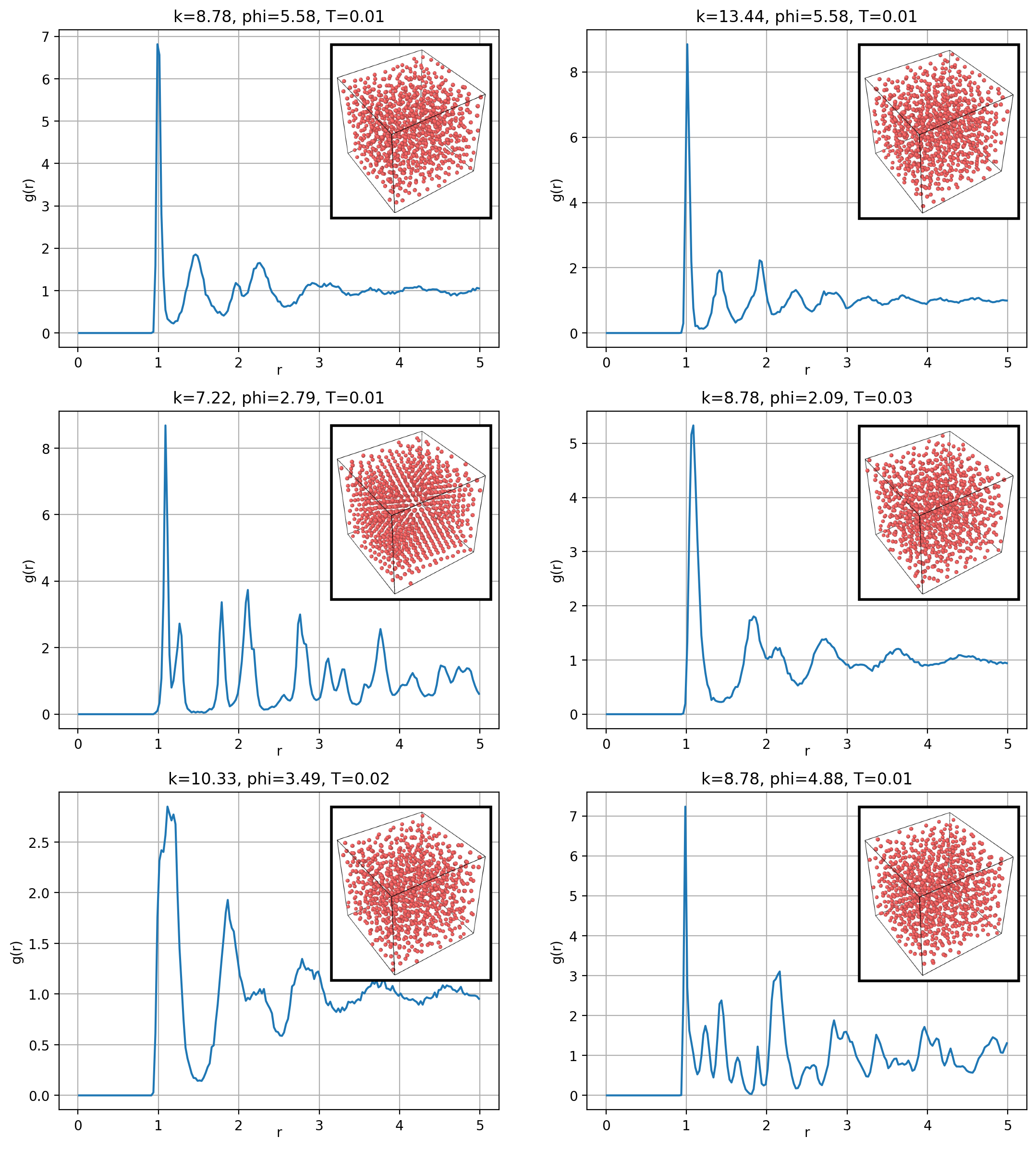}
    \caption{A sample gallery of radial distribution function (RDF) results and particle visualizations from the generated dataset. $k$ and $\phi$ are parameters of the oscillating pair potential, while $T$ is dimensionless temperature}
    \label{fig:gallery}
\end{figure}


\pagebreak
\section{Shape generation demonstration}

To sanity-check our model and the baseline, we start by investigating two shape-generation tasks, which is a more natural task for Diffusion with Transolver, and typical of point-cloud generative models. We train each model to generate a unit sphere and simple torus from normally-distributed points by training them on points sampled uniformly on the surfaces of these shapes. The results are shown in Tab.~\ref{tab:shape-results}. (Note that due to the translation-invariance of MDDM, we have re-centered the points after each diffusion step during sampling.) Sinkhorn loss, a metric for point-cloud matching, is reported.

\begin{table}[H]
    \renewcommand{\arraystretch}{1.1}
    \centering
    \caption{Sinkhorn loss for the sphere and torus generated by our model and the baseline}
\begin{tabular}{R{2.7cm}||C{2.5cm}|C{2.5cm}} \hline
    \vspace{3pt}\textbf{Model} & \textbf{Sinkhorn Loss Sphere} & \textbf{Sinkhorn Loss Torus}\\ \hline
        Diff. w/ Transolver & \textbf{0.0746} & \textbf{0.0155} \\
        \textit{MDDM (Ours)} & 0.0796 & 0.0165 \\
        \hline \end{tabular}
    \label{tab:shape-results}
\end{table}

\begin{figure}[H]
    \centering
    \includegraphics[width=0.6\textwidth]{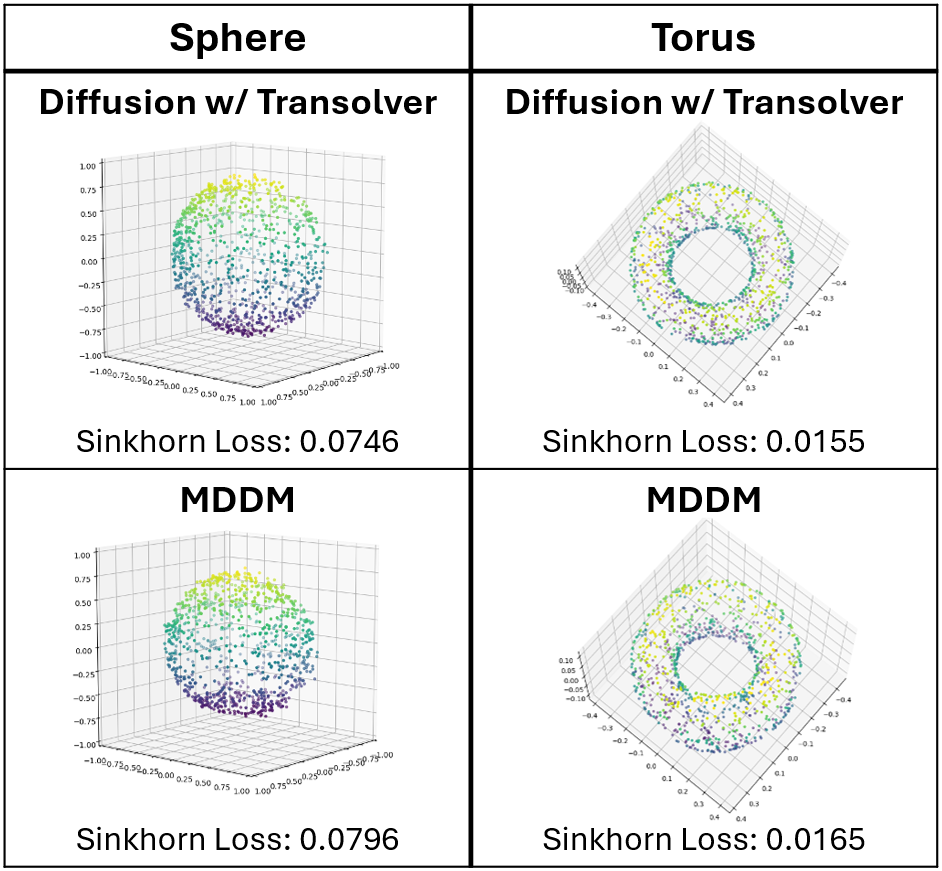}
    \caption{Sphere- and torus-matching results for our method and the baseline.}
    \label{fig:shapes}
\end{figure}

Figure \ref{fig:shapes} contains renders of the point clouds generated by each method, which were qualitatively successful. Table \ref{tab:shape-results} reveals that our model performs nearly as well as the baseline for shape generation tasks. This indicates that both our model and the baseline are working as intended and can be trained on the MD task.

\section{Code listings} \label{sec:listings}

This section contains Python code for simplified versions of the modules used for the denoiser in MDDM. \texttt{PeriodicBox()} is a utility for handling periodic boundary conditions, while \texttt{PBCNet()} is a module that contains several PBCConv layers, as defined in Fig.~\ref{fig:denoiser} and Eqn.~\ref{eqn:pbc-conv}. Each of these are showcased below.

\subsection{\texttt{PeriodicBox()}}
This module provides several utilities for applying periodic boundaries, which is necessary for dealing with systems of particles. It contains the method \texttt{wrap\_nearest()} for computing the shortest vector from one particle to another, wrapping across the periodic boundaries as necessary.  \texttt{wrap\_within()} wraps a particle's position back within the periodic box, for example if noising had removed it. Methods \texttt{knn\_tensor()} and \texttt{knn\_edges()} enable computation of a k-nearest neighbor graph under the presence of PBCs. The \texttt{PeriodicBox()} module is used extensively throughout the diffusion model, to wrap across periodic boundaries and to determine nearest neighbors in the periodic domain. 

\begin{lstlisting}[style=python,caption={PeriodicBox module, for handling wrapping around periodic boundaries, including creating of k-nearest neighbors graph},label={lst:periodicbox}]
class PeriodicBox(nn.Module):
    def __init__(self, Ls = torch.tensor([1., 1., 1.])):
        super().__init__()
        if type(Ls) != torch.tensor:
            Ls = torch.tensor(Ls, dtype=torch.float)
        self.L = Ls
    def wrap_nearest(self, A, B):  # Get vector from A to B, wrapping as necessary
        dr = B - A
        return dr + self.L*(dr < -self.L/2) - self.L*(dr > self.L/2)
    def wrap_within(self, X):      # Wrap points X to be within the periodic box
        return X - self.L * (X > self.L) + self.L * (X < 0)
    def wrap_dist(self, A, B):     # Get distance from A to B, wrapping as necessary
        return torch.norm(self.wrap_nearest(A, B), dim = -1)
    def knn_tensor(self, pts, k):
        D = self.wrap_dist(pts[:, None, :], pts[None, :, :])
        nearest = torch.topk(D, k+1, dim=1, largest=False)[1][:,1:]
        return nearest
    def knn_edges(self, pts, k):
        nearest = self.knn_tensor(pts, k)
        src = torch.arange(nearest.size(0)).repeat_interleave(nearest.size(1))
        dest = nearest.flatten()
        edges = torch.stack((src, dest), dim=1).long()
        return edges
\end{lstlisting}

\subsection{\texttt{PBCNet()}}
\texttt{PBCNet()} is a PyTorch module that serves as the denoising model in MDDM. The forward pass takes particle coordinates as input, computes the k-nearest neighbor graph, and then performs several periodic graph convolutions. Finally, each particle's convolved features are passed through an MLP to compute an estimate of what noise was added to the data to arrive at the input. See Fig.~\ref{fig:denoiser} for the network structure and Eqn.~\ref{eqn:pbc-conv} for the graph convolution formula.

\begin{lstlisting}[style=python,caption={PBCNet module, a graph neural network architecture in which nodes exchange absolute-position-invariant information with nearest neighbors, including across periodic boundaries},label={lst:pbcnet}]
class PBCNet(nn.Module):
    def __init__(self, n_dims, n_features, n_conv_features, n_out,
                 L=10, k=12, n_conv_layers=2,
                 conv_mlp_size=[64,64], out_mlp_size=[128,128],
                 periodic_box=None, residual=True, concat_features=True):
        super().__init__()
        self.n_dims, self.n_features = n_dims, n_features
        self.n_conv_features, self.n_out = n_conv_features, n_out
        self.k, self.n_conv_layers = k, n_conv_layers
        self.residual, self.concat_features = residual, concat_features
        self.pbc = periodic_box or PeriodicBox([L]*n_dims)

        self.in_conv = MLP([n_dims+n_features, *conv_mlp_size, n_conv_features])
        self.convs = nn.ModuleList([
            MLP([n_dims + 2*n_conv_features + (n_features if concat_features else 0),
                 *conv_mlp_size, n_conv_features])
            for _ in range(n_conv_layers)
        ])
        self.out_mlp = MLP([n_conv_features + (n_features if concat_features else 0),
                            *out_mlp_size, n_out])

    def get_edges(self, pts):
        """Return neighbor edges and wrapped displacement vectors."""
        edges = self.pbc.knn_edges(pts, self.k)
        vecs = self.pbc.wrap_nearest(pts[edges[:,0]], pts[edges[:,1]])
        return edges, vecs

    def conv(self, mlp, vecs, edges, node_f=None, shared_f=None):
        """Apply one convolution layer and pool edge features."""
        feature_vecs = [vecs]
        if node_f is not None:
            src, dst = node_f[edges[:,0]], node_f[edges[:,1]]
            feature_vecs += [src, src - dst]
        if shared_f is not None:
            feature_vecs.append(shared_f[edges[:,0]])
        A = torch.cat(feats, -1).float()
        B = mlp(A).view(node_f.shape[0], self.k, self.n_conv_features)
        return torch.max(B, 1)[0]

    def forward(self, X):
        # This implementation does not allow a batch dimension
        coords, shared = X[:,:self.n_dims], X[:,self.n_dims:]
        edges, vecs = self.get_edges(coords)

        # Initial PBC-Conv layer
        f = self.conv(self.in_conv, vecs, edges, shared_f=shared)
        shared_f = shared if self.concat_features else None

        # Subsequent PBC-Conv layers, optionally residual
        for mlp in self.convs:
            f_new = self.conv(mlp, vecs, edges, node_f=f, shared_f=shared_f)
            f = f + f_new if self.residual else f_new

        # Output particle-wise MLP
        if self.concat_features: f = torch.cat([shared_f, f], -1)
        return self.out_mlp(f)
\end{lstlisting}

\subsection{MD Input Script}
To generate a large dataset of MD results, we define the following string in a Python script. This string's bracketed values are replaced using \texttt{.format()} and the result is written to a LAMMPS \cite{LAMMPS} input script file. 1000 such scripts are generated and simulated in LAMMPS to generate the dataset of bulk conformations. Each simulation anneals to the target temperature from 10 times the (absolute) target temperature over the course of 100,000 time steps in NVT, and then allows further equilibration for an additional 100,000 time steps. The dimensionless time step is set as 0.005. Particles are initialized in a 10x10x10 simple cubic lattice, perturbed slightly with uniform noise, and assigned random velocity vectors sampled from a standard normal distribution.

\begin{lstlisting}[style=python, caption={LAMMPS input script for running a 1000-particle MD self-assembly with a custom tabulated potential},label={lst:mdscript}]
IN_CUSTOM = """
units lj
atom_style atomic
dimension 3
boundary p p p

region box block 0 10 0 10 0 10
create_box 1 box

lattice sc 1.0
create_atoms 1 box

pair_style table linear 1000
pair_coeff 1 1 custom.table CUSTOM

mass 1 1.0

velocity all create 1.0 12345 dist gaussian
displace_atoms all random 0.1 0.1 0.1 12345

neighbor 0.3 bin
neigh_modify delay 0 every 1 check yes

# Nose-Hoover Thermostat for NVT
fix 1 all nvt temp {Ti} {Tf} $(100.0*dt)

timestep 0.005
thermo 100
thermo_style custom step temp pe ke etotal

run {num_steps}
dump 1 all atom {dump_every} {dump_file}
run {num_steps}
"""
\end{lstlisting}


\section{Mathematical claims}\label{sec:proofs}

\subsection{Uniformity of a wrapped Gaussian}
\begin{theorem}
A normal distribution with standard deviation $L$ that has been wrapped by a periodic boundary function onto the interval $[0,L)$ approximates a uniform distribution on $[0,L)$.
\end{theorem}

\begin{proof}  
(Note: This proof does not quantify the error in the approximate result. However, our experiments have shown that the probability density function of a wrapped standard normal distribution is virtually indistinguishable from that of a uniform distribution at all points; the following proof explains why.)

We start by showing that the theorem holds for the standard normal distribution with mean 0 and standard deviation 1, and without loss of generality we assert that the result extends to other means and standard deviations.

Let $X$ be a random variable with an approximately standard normal distribution. According to the Central Limit Theorem, a normal distribution can be closely approximated by an Irwin-Hall distribution with $N=12$ (that is, the distribution describing the sum of 12 independent uniform random variables on $[0,1)$, subtracting 6 in this case to re-center at zero), which has mean 0 and standard deviation 1 when expressed using the PDF below, adapted from Marengo et al.~\cite{irwin-marengo-2017}:
\begin{align*}
X &\overset{\cdot}{\sim} N(0, 1), \quad \frac{1}{\sqrt{2\pi}} e^{-\frac{x^2}{2}} \approx  f_X(x) = \frac{1}{2(12-1)!}\sum_{r=0}^{12} (-1)^r \binom{12}{r} \text{sgn}(x+6-r)(x+6-r)^{12-1} 
\end{align*}

Suppose $Y=g_L(X)$ is a periodic boundary function of $X$, where $g_L(a)$ is defined as $g_L(a) = a + k_a L$, for the unique integer $k_a$ such that $0 \leq g_L(a) < L$. We aim to derive the PDF of $Y$. For a general non-monotonic transformation, the PDF of $Y$ is given by the change of variables:
\begin{align*}
f_Y(y) = \sum_{x:\, g_L(x) = y} \frac{f_X(x)}{|g_L'(x)|}.
\end{align*}

Note that this sum accounts for all possible preimages $\forall k \in \mathds{Z}:\; x = y + k L$, but $f_Y(y)$ is only nonzero on the interval $[0,1)$, and only $k$ values from -6 to 5 result in nonzero probability density. Also, $|g_L'(x)|$ evaluates to 1 everywhere. Thus, the sum is:

\begin{align*}
f_Y(y) = 
\begin{cases} 
{\sum_{k=-6}^{5} \frac{1}{2(12-1)!} \sum_{r=0}^{12} (-1)^r \binom{12}{r} \text{sgn}(y+k+6-r)(y+k+6-r)^{12-1}} & \text{if } 0 \leq y < 1, \\
0 & \text{otherwise.}
\end{cases}
\end{align*}

For y values on the interval $[0,1)$, this expands to:
\begin{flalign*}
f_Y(y)\Big|_{0\leq y < 1} &= \Big(-462 \left(y-6\right)^{11}+\left(y-1\right)^{11}+11 \left(y+1\right)^{11}-55 \left(y+2\right)^{11}+165 \left(y+3\right)^{11} \\
&\qquad +330 \left(y-5\right)^{11}-55 \left(y-10\right)^{11}-11 \left(y-2\right)^{11}+11 \left(y-11\right)^{11}+55 \left(y-3\right)^{11} \\
&\qquad -165 \left(y+8\right)^{11}+55 \left(y+9\right)^{11}-11 \left(y+10\right)^{11}+\left(y+11\right)^{11}+462 \left(y-7\right)^{11} \\
&\qquad -330 \left(y-8\right)^{11}-165 \left(y-4\right)^{11}+165 \left(y-9\right)^{11}-330 \left(y+4\right)^{11}+462 \left(y+5\right)^{11} \\
&\qquad -462 \left(y+6\right)^{11}+330 \left(y+7\right)^{11}-\left(y-12\right)^{11}-y^{11}\Big) \Big/ \big( 2\; (12-1)! \big)\\
&= 1
\end{flalign*}

Through symbolic manipulation, the above expression evaluates to exactly 1. Therefore, the PDF is simply:
\begin{align*}
f_Y(y) = 
\begin{cases} 
1 & \text{if } 0 \leq y < 1, \\
0 & \text{otherwise.}
\end{cases}
\end{align*}

This is recognized as a uniform distribution on $[0,1)$. Therefore, because the wrapped Irwin-Hall distribution is exactly uniform, we conclude the distribution $\Nc(0,1)$ wrapped onto the interval $[0,1)$ closely approximates the uniform distribution $\Uc[0,1)$.

Without loss of generality, this relationship holds for any mean $\mu$ and any standard deviation $L$ by rescaling the distributions and shifting the periodic region. Thus, the distribution $\Nc(\mu,L)$ wrapped onto the interval $[0,L)$ approximates the uniform distribution $\Uc[0,L)$.

\qedhere
\end{proof}

\subsection{Variational Inference for a reinterpretation of forward diffusion}

\begin{theorem}
For a forward diffusion process defined by $\xv_{t} = \xv_0 + \sqrt{\alpha_t}\epsilonv;\; \epsilonv \sim \Nc(\boldsymbol{0}, \boldsymbol{I})$, the posterior mean for variational inference is $\mu_t = \frac{x_t-x_0}{\alpha_t}\alpha_{t-1}+x_0$ and the posterior variance is $\sigma_t^2 = \frac{\alpha_{t-1}(\alpha_t-\alpha_{t-1})}{2\alpha_t}$.
\end{theorem}
\begin{proof}
To start, we know the ELBO of the diffusion term \cite{luo2022understanding} is 
\begin{align}
    \Ev_{q(x_{1}:x_{T}|x_{0})}\log{\frac{p(x_{0}:x_{T})}{q(x_1:x_T|x_0)}} 
    = & \, \Ev_{q(x_1|x_0)}\log P_{\theta}(x_0|x_1) 
    - D_{KL}(q(x_1|x_0)||p(x_T)) \notag \\
    & - \sum\limits_{t=2}^{T} \Ev_{q(x_1|x_0)} 
    \big(D_{KL}(q(x_{t-1}|x_t, x_0)||p_\theta(x_{t-1}|x_t))\big)
\end{align}

For variational inference, we are interested in deriving a mathematical expression for the third term of this equation, the \textit{denoising matching term}. In particular, we assume that $q$ follows a Gaussian distribution, and we want to minimize the KL divergence. Therefore, the goal is to find the mean and variance of this distribution that would result in the closest possible match between distributions $p$ and $q$. To do this, we first rewrite $x_t$ and $x_{t-1}$ in terms of our denoising process from \ref{eqn:fwd-noise} as follows:
\begin{flalign}
    x_t \quad &= x_0 + \sqrt{\alpha_t}\epsilonv \notag \\
    x_{t-1}   &= x_0 + \sqrt{\alpha_{t-1}}\epsilonv \notag \\
    x_t \quad &= x_{t-1} + (\sqrt{\alpha_t}-\sqrt{\alpha_{t-1}})\epsilonv,
\end{flalign}

where $\epsilonv \sim \Nc(\boldsymbol{0}, \boldsymbol{I})$. Next, we look to Bayes' Rule to obtain a relationship between the conditional probabilities:
\begin{equation}
    q(x_{t-1}|x_t, x_0) = \frac{q(x_t|x_{t-1}, x_0)\; q(x_{t-1}|x_0)}{q(x_t|x_0)}
\end{equation}

Using the above equations, we can determine proportionality relationships in the denoising matching term:
\begin{flalign}
    &\quad
    \frac{\Nc(x_t;\; x_{t-1},\; (\alpha_t-\alpha_{t-1})\boldsymbol{I}) \cdot \Nc(x_{t-1};\; x_0,\; \alpha_{t-1}\boldsymbol{I})}{\Nc(x_t;\; x_0,\; \alpha_t\boldsymbol{I})}\dots
    \notag \\&\qquad
    \propto \exp\left(-\left[ \frac{(x_t-x_{t-1})^2}{\alpha_t - \alpha_{t-1}} + \frac{{(x_{t-1} - x_0})^2}{\alpha_{t-1}} - \frac{(x_{t}-x_0)^2}{\alpha_t} \right]\right) &
    \notag \\&\qquad
    \propto \exp \left(\frac{-1}{2}\left[\frac{2\alpha_t^2}{\alpha_t\alpha_{t-1}\left(\alpha_t-\alpha_{t-1}\right)}\right]
    \left[x_{t-1}^2 -2x_{t-1}\left(\frac{x_t-x_0}{\alpha_t}\alpha_{t-1}+x_0\right)+\left(\frac{x_t-x_0}{\alpha_t}\alpha_{t-1}+x_0\right)^2\right]
    \right)
    \notag \\&\qquad
    \propto \Nc\left(x_{t-1};\; \mu = \frac{x_t-x_0}{\alpha_t}\alpha_{t-1}+x_0,\; \sigma^2 = \frac{\alpha_{t-1}(\alpha_t-\alpha_{t-1})}{2\alpha_t}\right) \label{eqn-mean-var}
\end{flalign}

Equation \ref{eqn-mean-var} therefore gives a formulation for mean and variance under our modified noising strategy. Note that to maintain training robustness, we regulate the training objective to always be distributed as $\Nc(\boldsymbol{0}, \boldsymbol{I})$. Therefore, during sampling, the mean becomes:

\begin{equation}\label{eqn:mean}
    \mu = \frac{x_t-\hat{x}_0}{\alpha_t}\alpha_{t-1}+\hat{x}_0 = ( \frac{\alpha_{t-1}}{\alpha_t}-1)\cdot \left(\sqrt{\alpha_t}\; \boldsymbol{\epsilon}_\theta (\xv_t, t)\right) + \xv_t
\end{equation}

The mean and standard deviation in Eqns.~\ref{eqn-mean-var} and \ref{eqn:mean} are then used during sampling (Algorithm \ref{alg:sample}).

\qedhere
\end{proof}


\end{document}